\pdfoutput=1

\documentclass[11pt]{article}
\usepackage{graphicx}
\usepackage{multirow}
\usepackage{amsmath}
\usepackage{bm}
\usepackage{adjustbox}
\usepackage{booktabs,siunitx}
\usepackage{amssymb}

\usepackage[]{acl}

\usepackage{times}
\usepackage{latexsym}

\usepackage[T1]{fontenc}

\usepackage[utf8]{inputenc}

\usepackage{microtype}

\title{Lack of Fluency is Hurting Your Translation Model}

\author{Jaehyo Yoo \\
    Korea University \\
  \texttt{jaehyoyoo@korea.ac.kr} \\\And
  Jaewoo Kang \\
  Korea University \\
\texttt{kangj@korea.ac.kr} \\}

\begin{document}
\maketitle
\begin{abstract}
Many machine translation models are trained on bilingual corpus, which consist of aligned sentence pairs from two different languages with same semantic. However, there is a qualitative discrepancy between train and test set in bilingual corpus. While the most train sentences are created via automatic techniques such as crawling and sentence-alignment methods, the test sentences are annotated with the consideration of fluency by human. We suppose this discrepancy in training corpus will yield performance drop of translation model. In this work, we define \textit{fluency noise} to determine which parts of train sentences cause them to seem unnatural. We show that \textit{fluency noise} can be detected by simple gradient-based method with pre-trained classifier. By removing \textit{fluency noise} in train sentences, our final model outperforms the baseline on WMT-14 DE$\rightarrow$EN and RU$\rightarrow$EN. We also show the compatibility with back-translation augmentation, which has been commonly used to improve the fluency of the translation model. At last, the qualitative analysis of \textit{fluency noise} provides the insight of what points we should focus on.
\end{abstract}

\section{Introduction}

Machine translation is a task of generating translated sentences from given language to other. The final goal of the task is to create the fluent sentences which are hard to distinguish from golden translations. Since the translation model needs huge language prior, many studies utilized the specific pipeline to generate the large bilingual benchmark through crawling, sentence-alignment, and many other filtering methods \cite{europarl-2005, cettolo-etal-2012-wit3, banon-etal-2020-paracrawl}. However, these automatically produced benchmark may not completely follow our intrinsic desire since we usually evaluate our translation model on human-annotated sentences, whose quality is guaranteed by human evaluation. Since the cost of human evaluation on the entire training data is extremely expensive, we cannot be convinced that the discrepancy between train and test distribution would not hurt our translation model.

\begin{figure}[t]
\begin{center}
\includegraphics[width=0.98\linewidth]{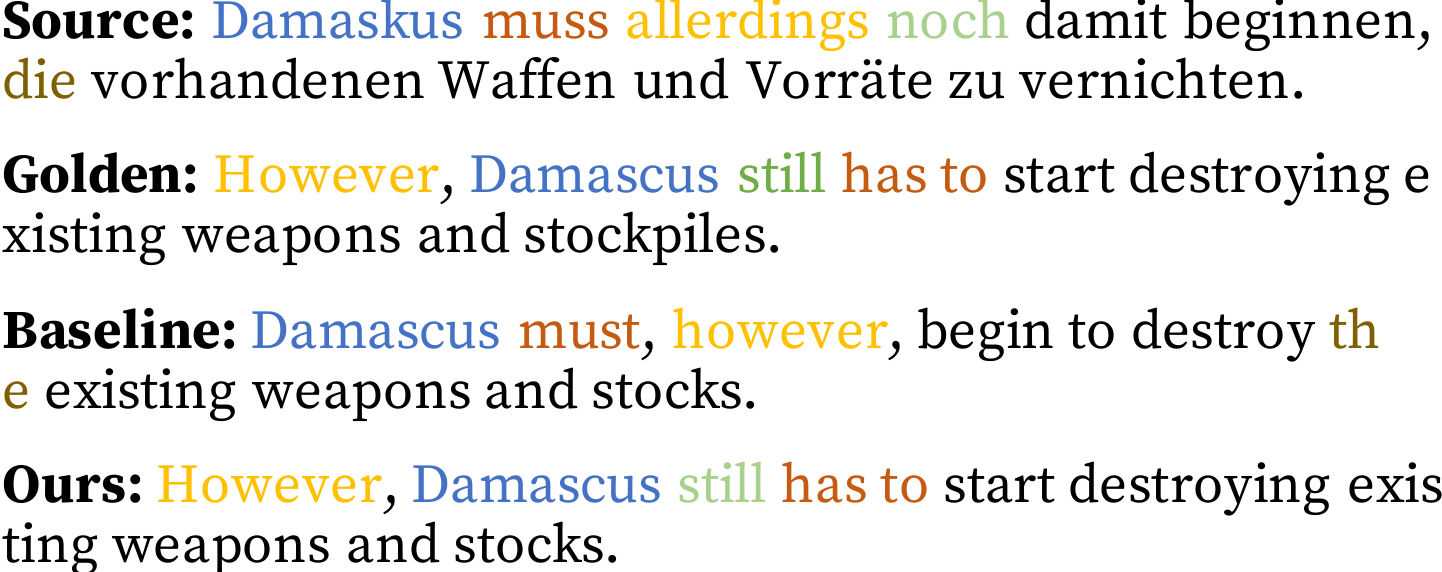}
\end{center}

\caption{An example from WMT-14 DE$\rightarrow$EN test set. We translate the given source sentence with the baseline and our final model for each, then compare them to the golden answer. For convenience, we marked the tokens with same color if their meanings are equal.}
\label{fig:long}
\label{fig:onecol}
\label{fig:intro}
\end{figure}

We suppose this discrepancy appears by the strict and inflexible alignment of sentences, causing the lack of fluency to processed train data. Since the most heuristic rules are not perfect, the processed data could suffer from the alignment error issue \citep{goutte-etal-2012-impact}. Although the alignment is done successfully, the translation of given sentence may seem awkward since the purpose of translation in the most websites focus on the transfer of semantic, not its naturalness.

Since numerous factors could be the reason of fluency lack, it is hard to define the concept of naturalness. In Figure \ref{fig:intro}, for example, we compare the translation of baseline model to its human-annotated golden answer. As we can see, the baseline model gets low BLEU score for too strict alignment of word-order, simplification by omission (for \textit{noch}), and explicitation of determiner (for \textit{die}). Nonetheless, we can't say that this translation is wrong due to its correct factual consistency. Since the standard of evaluation is ambiguous and different for each individual annotator, human evaluation also suffers from the difficulty of measuring the fluency without any guideline \citep{clark-etal-2021-thats}. Then how can we define the range of unnatural points in the given sentence without human evaluation?

One of the solution is to use the language model prior, which is the probability of generating the next word without seeing the source sentence for translation. \citet{baziotis-etal-2020-language} defines Kullback-Leibler divergence using language model prior to regularize the posterior of translation model on low-resource setting. On the other hand, \citet{miao-etal-2021-prevent} claims that the language model prior could be overconfident due to the excessive attention towards the partial translation. However, these methods are still abstract to catch the explicit evidence distracting the fluency.

In our work, we utilize the prior of monolingual data to determine whether the given sentence is native-like or not. We first train the classifier to check if it is easy to distinguish train or test sentences from bilingual corpus and sentences from monolingual corpus. From the experiment, we find that human-annotated test set is much harder to classify from monolingual corpus, while the classifier distinguishes the train set with much higher accuracy.

Recent work \citep{seq_clf} examined that a few tokens which play an important role in sequence classification problems can be found by gradient-based methods. Following the previous work, we show that we can mark which parts of the given target sentence seem awkward compared to monolingual texts. We define these marked tokens as \textit{fluency noise}. Eventually, we mask \textit{fluency noise} during train time, and outperform the baseline model by up to 1.0 BLEU score. We validate the compatibility with the back translation augmentation method \citep{sennrich-etal-2016-improving}, since the method has also been used to improve the fluency of the model. We lastly analyze \textit{fluency noise} compared to frequently appeared phrases in train corpus using simple statistics. Through the analysis, we show that \textit{fluency noise} can not be defined by just frequentist approach.

Our main contributions are summarized as follows: 
\begin{itemize}
    \item{
        We directly show the existence of \textit{fluency noise} in the translation benchmark against the monolingual corpus.
    }
    \item{
        Without any additional parameter or data augmentation, we outperform the baseline translation model by just masking the input sentences partially.
    }
    \item{
        We provides the qualitative analysis of \textit{fluency noise} in our final model. We expect the analysis can be a fine milestone to the studies in data-centric translation field.
    }
\end{itemize}

\section{Related Works}
Visualizing and understanding the composition of deep learning model has been a big challenge regardless of its field. In NLP, \citet{li-etal-2016-visualizing} tried to explain the importance of each word embedding with its gradient. Layer-wise relevance propagation (LRP) \citep{lrpuse}, which is a measure to compute the contribution of neurons, is also a good choice to analyze the composition of the model \citep{voita-etal-2019-analyzing}. Recently, \citet{seq_clf} proves that the first-derivatives or LRP of the word embedding are both excellent to extract the important words in sequence classification problem using BERT \citep{devlin-etal-2019-bert}. Following the work, we extract \textit{fleuncy noise} from traind classifier by just calculating the norm of gradient for each token. 

Back-translation augmentation \citep{sennrich-etal-2016-improving} is a popular method utilizing the monolingual data as the target sentences of translation model. Since the decoder could learn to generate the monolingual sentences during the training, the method has been used in many studies \cite{edunov-etal-2018-understanding, hoang-etal-2018-iterative, caswell-etal-2019-tagged} to compare or improve the fluency of the model. \citet{cai-etal-2021-neural} also used the monolingual corpus as its translation memory to retrieve the aligned target sentences. In our work, we only utilize the monolingual data to modify the train data for masking, and detach it during the translation process. We also show that our proposed method is complementary to back-translation.

\begin{figure*}[t]
\begin{center}
\includegraphics[width=0.99\textwidth]{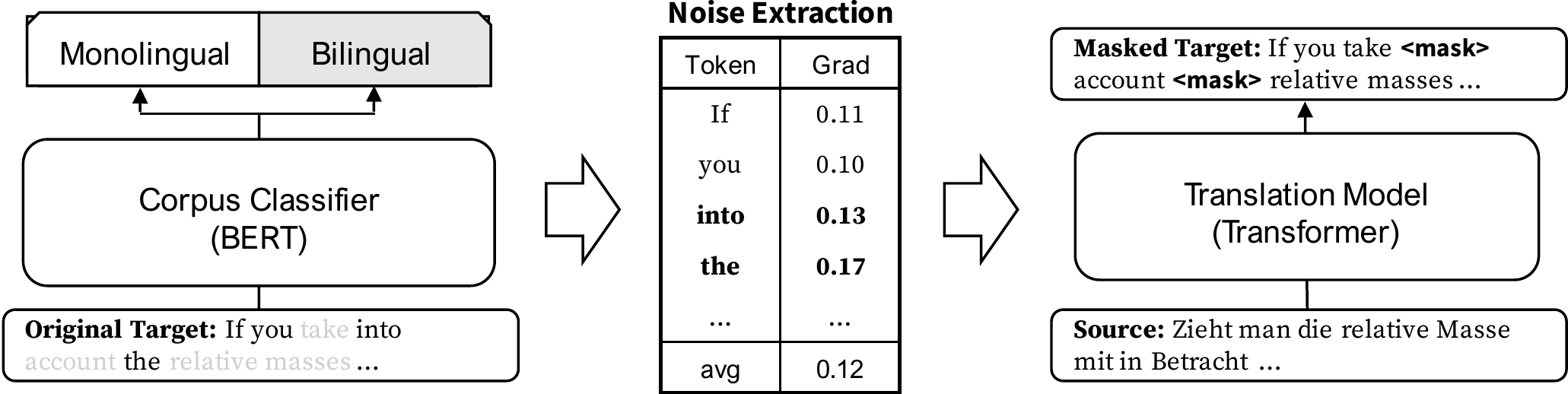}
\end{center}

\caption{Overall framework for masking the given target sentence. The corpus classifier is firstly trained on English sentences from both monolingual and bilingual corpus. After the training is finished, we could get the gradient norm of each token embedding in the target sentence as regard to the output logit of the classifier. During the inference time, we mask the content words and get the logit by only attending to function words. Lastly, we mask the tokens whose gradient norm is over the average and use the masked target for translation task instead of the original target. }
\label{fig:pipeline}
\end{figure*}

Translationese can also be the topic related to our works. It refers to the remaining artifact caused by translation process, including lexical difference and more explicit or simpler representation \citep{baker}. There has been challenges to detect translationese by distinguishing the translated sentences \citep{kurokawa-etal-2009-automatic}, or predicting the source language from given translations \citep{koppel-ordan-2011-translationese}.
Following the previous works, \citet{pylypenko-etal-2021-comparing} showed that BERT could catch the feature of translationese with high correlation to the traditional hand-crafted features. \citet{riley-etal-2020-translationese} also trained the translationese classifier to tag the sentences in training data to control the output translation. Our work is quite similar to the detection of translationese. However, we exclude the content words(adjectives, adverbs, nouns, and verbs) for masking since the deletion of content words could cause the hallucination problem \citep{raunak-etal-2021-curious}. We only focus on syntactic and function words in experiments, without the consideration of lexical objects.

\section{Experimental Setup}
\subsection{Data}
We mainly conduct our experiments on WMT-14 German$\rightarrow$English (DE$\rightarrow$EN), and WMT-14 Russian$\rightarrow$English (RU$\rightarrow$EN) translation benchmarks. The dataset consists of 4.5M sentence pairs for DE$\rightarrow$EN, and 2.5M sentence pairs for RU$\rightarrow$EN.

For classification, we use the entire English sentences of bilingual corpus by dividing 5K sentences for dev, and another 5K sentences for test. Then, we sample the monolingual English sentences with the same ratio from News Crawl 2014. The number of tokens of each sampled sentence is set to be equal to each sentence's in bilingual corpus to avoid the length bias.

We also collect human-annotated test set generally used in translation task to compare the difficulty of classification. The collection consists of 14K English sentences from newstest2009-2013. We divide this set by each 1K sentences for dev and test, and use the same strategy to build the benchmark for classification.

Lastly, we evaluate our translation models on newstest2013 for dev, and newstest2014 for test.

\subsection{Corpus Classification}

Training of the classifier is proceeded by binary classification, predicting whether the given sentence comes from monolingual corpus or bilingual corpus. For the use of huge linguistic prior, we select BERT \citep{devlin-etal-2019-bert} for our classification model. Similar to the works using BERT for sequence classification \citep{pylypenko-etal-2021-comparing}, the embedding of [CLS] token in last layer is pooled and fed into linear projection. Since our final goal is to find and mask the awkward tokens in the training data for translation, we have a concern that masking content words would yield the misalignment of semantic, which can cause the hallucination problem \citep{raunak-etal-2021-curious}. So we avoid the problem by masking the content words\footnote{In our work, we define noun, verb, adjective, and adverb as the content words. Others are defined as the function words.} during the inference of classifier. We have an expectation that the model can distinguish the native-likeliness with only function words and its position since they learn the semantic role during train time. The equation of inference is summarized as follows :

\begin{equation}
\label{equation:classification}
    \begin{aligned}
    & L_t^m, \: L_t^b = f(X_t, C_t)
    \end{aligned}
\end{equation}

$X_t$ refers to the token sequence of the target sentence, and $C_t$ refers to the content words in $X_t$. The classifier $f$ masks $C_t$ during the inference and calculates two output logits, $L_t^m$ and $L_t^b$ for the prediction of monolingual and bilingual corpus. We will use only $L_t^b$ for the extraction of noise in next section.

\subsection{Noise Extraction}

Following the prior work by \citet{seq_clf}, we can sort the input tokens by their contributions to the output classification logit. Namely, the words with high contribution to the prediction of bilingual text refer to the words seem awkward compared to the monolingual text. We calculate the contribution score by getting the first derivative of token embedding for bilingual logit $L_t^b$, and taking the L2 norm on it. We denotes $G_t = \{x \mid x \in X_t \text{ and } x \notin C_t \}$ is the candidate set consisting of tokens which are substantially used for classification. Then, our selection is defined as follows:

\begin{equation}
\label{equation:noise_extraction}
    \begin{aligned}
    F_t = \{x \mid x \in G_t \text{  and  } \lVert {\partial{L_t^b} \over \partial{x}} \rVert \ge \bar{g_t} \} \\
    where \quad \bar{g_t} = {1 \over N} \sum_{x \in G_t} \lVert {\partial{L_t^b} \over \partial{x}} \rVert
    \end{aligned}
\end{equation}

We define \textit{fluency noise}, which are the tokens distracting the given sentence from being native-like, by denoting it as $F_t$. We only select the tokens with higher gradient norm than its average to be included in $F_t$. We exclud the symbol tokens for calculation such as [CLS] or [SEP].

\subsection{Masked Translation}
\label{sssec:masked_translation}
Finally, our translation model is trained with masked target sentence by replacing the tokens in $F_t$ to <mask> symbol. The model doesn't learn to generate or attend to <mask> during train time. We also consider that \textit{fluency noise} can be not reliable if the corpus classifier lacks the confidence on its result. So we set the confidence threshold $\gamma$ ranging from 0 to 1, to mask only the sentences with higher classification likelihood than $\gamma$. The likelihood is calculated via applying softmax to Equation~\eqref{equation:classification}. We evaluated our final models using SacreBLEU \cite{papineni-etal-2002-bleu, post-2018-call}, with identification string on footnote\footnote{nrefs:1|case:mixed|eff:no|tok:13a|smooth:exp|version:2.0.0"}.

\subsection{Implementation Details}

For corpus classification, we use cased BERT-base model with a batch size of 128 on WMT-14 DE$\rightarrow$EN \& RU$\rightarrow$EN, and a small batch size of 32 on newstest. We trained the large sets with 3 epochs, and the small sets with 5 epochs. Since we does not consider the content words, we excludes Wiki-Headlines data in WMT-14 RU$\rightarrow$EN, which is the set consisting of phrase translations. 

We use Transformer-Base \citep{vaswani} architecture for our translation model, implemented in OpenNMT-py\footnote{https://github.com/OpenNMT/OpenNMT-py}. Byte-pair encodings (BPE) \citep{sennrich-etal-2016-neural} is used as our tokenization strategy with 32K vocab size, sharing the source and target dictionary. Following the previous work \citep{sennrich-etal-2016-neural}, we transliterate Russian texts into Latin with ISO-9. 

Note that since our methodology is model-agnostic, we can use other recent models as our classifier \citep{liu2019roberta} or translator \cite{2019-t5, liu-etal-2020-multilingual-denoising}.

\section{Results}

\subsection{Corpus Classification}
\label{sssec:corpus_clf_result}

Since the performance of corpus classification determines the difficulty of distinguishing the sentences from bilingual corpus, we could directly compare the train (DE$\rightarrow$EN or RU$\rightarrow$EN) set to the human-annotated test (newstest) set of translation task. Table~\ref{tab:corpus_clf} shows the accuracy of corpus classifier. The result shows that sentences from bilingual corpus are quite distinguishable, which is complementary to prior works of translationese \cite{koppel-ordan-2011-translationese, pylypenko-etal-2021-comparing}. For fair comparison, we also conduct the experiment on the shrunk set of DE$\rightarrow$EN and RU$\rightarrow$EN of which the size is equal to newstest. We could check that the shrunk settings outperform the model trained on newstest by 4.8-7.7\%. In other words, the sentences in newstest are much harder to distinguish from monolingual texts although the experimental setting is equal.

\begin{table}[]
\centering
\resizebox{\columnwidth}{!}{%
\begin{tabular}{clcc}
\toprule
                  From     & Trained on  & \# sents  & Accuracy \\
\midrule                    
\multirow{2}{*}{DE$\rightarrow$EN} & Full  & 8,997,570 & \textbf{95.8}     \\
                       & Small  & 24,040 & 90.1     \\
\midrule                       
\multirow{2}{*}{RU$\rightarrow$EN} & Full & 4,069,180 & \textbf{92.0}     \\
                       & Small & 24,040    & 87.2          \\
\midrule                       
\multirow{1}{*}{newstest} & Full & 24,040 & 82.4          \\
\bottomrule                       
\end{tabular}%
}

\caption{
Accuracy of corpus classifier. Each train set consists of mixed sentences from bilingual corpus and monolingual corpus with the same ratio. \textbf{Full} : Full sentences used. \textbf{Small} : Number of sentences in the train data gets smaller to be equal to the size of newstest. 
}
\label{tab:corpus_clf}

\end{table}

\subsection{Masked Translation}
Our main results on translation task are reported in Table~\ref{tab:main_result}. As we discussed in Section~\ref{sssec:masked_translation}, the confidence threshold $\gamma$ is used to filter the sentences to mask \textit{fleuncy noise} or not. We search $\gamma$ in range of \{$0.5, 0.7, 0.8, 0.9, 1.0$\}. $\gamma=1.0$ means that no masking is used for translation, and we deal with this setting as our baseline. We derive the classification likelihood for each sentence from the classifier trained in Section~\ref{sssec:corpus_clf_result}.

For DE$\rightarrow$EN, we show that all masking strategies are effective to outperform the baseline. BLEU is generally proportional to $\gamma$ before it reaches to $1.0$. However, we confirm that excessive filtering is not helpful since the maximum score is acquired at $\gamma=0.8$. 

For RU$\rightarrow$EN, the model with $\gamma=0.5$ gets lower score than the baseline. Since the classification model is more calibrated than DE$\rightarrow$EN's, it leads to much higher ratio of masking. However, this degradation disappears by raising the threshold. Interestingly, the model keeps the score of its baseline despite the deletion of up to 8.8\% of entire tokens. The score reaches at its peak when $\gamma$ equals to $0.9$.

We could notice that the adjustment of $\gamma$ is required and the proper value depends on the confidence of trained classifier. Since finding $\gamma$ is somewhat exhaustive, it is worthy to define which parts of our benchmark is harmful to translation model. We also discuss about \textit{fluency noise} of our final models in Section~\ref{sssec:fleuncy_noise}.

\begin{table}[]
\centering
\begin{tabular}{ccccc}
\toprule
\multicolumn{1}{c}{\multirow{2}{*}{}} & \multicolumn{1}{c}{\multirow{2}{*}{\begin{tabular}[c]{@{}c@{}}$\gamma$\end{tabular}}} & \multicolumn{2}{c}{Mask Ratio} & \multirow{2}{*}{BLEU} \\
\multicolumn{1}{c}{}                  & \multicolumn{1}{c}{}                                                                                & Sentence        & Token        &                       \\
\midrule
\multirow{5}{*}{DE$\rightarrow$EN} & 1.0                                                            & 0.0 \%                                                          & 0.0 \%                                                       & 30.0 \\ 
                       & 0.9                                                            & 12.6 \%                                                         & 1.8 \%                                                      & 30.8 \\
                       & 0.8                                                            & 17.6 \%                                                         & 2.9 \%                                                      & \textbf{31.0} \\
                       & 0.7                                                            & 25.4 \%                                                         & 4.6 \%                                                      & 30.6 \\
                       & 0.5                                                            & 58.6 \%                                                         & 4.7 \%                                                      & 30.5 \\
\midrule
\multirow{5}{*}{RU$\rightarrow$EN} & 1.0                                                            & 0.0 \%                                                          & 0.0 \%                                                       & 32.2 \\
                       & 0.9                                                            & 17.1 \%                                                         & 3.6 \%                                                      & \textbf{33.2} \\
                       & 0.8                                                            & 28.6 \%                                                         & 6.3 \%                                                      & 32.6 \\
                       & 0.7                                                            & 41.2 \%                                                         & 8.8 \%                                                      & 32.2 \\
                       & 0.5                                                            & 71.8 \%                                                         & 13.2 \%                                                      & 31.5 \\
\bottomrule                      
\end{tabular}

\caption{
BLEU score for models trained on WMT-14 DE$\rightarrow$EN and RU$\rightarrow$EN. $\bm{\gamma}$ : confidence threshold defined in Section~\ref{sssec:masked_translation}. \textbf{Mask Ratio (Sentence)} : the ratio of sentences with higher classification likelihood than $\gamma$. \textbf{Mask Ratio (Token)} : the ratio of masked tokens to the entire tokens in train data.  
}
\label{tab:main_result}
\end{table}

\section{Analysis}

\subsection{Compatibility with Back Translation}
\label{sssec:back_translation}

Back-translation method \citep{sennrich-etal-2016-improving} deals with several points similar to our work. By utilizing the monolingual data as target sentences, it improves the general fluency of translation model. Our method is the deletion of \textit{fluency noise} which is the awkward point in given sentence compared to monolingual corpus. We can suppose if \textit{fluency noise} is far away from the distribution of monolingual prior, it can disturb the regularization of back-translation method and debiasing it would be helpful. 

In Table~\ref{tab:main_result}, we conduct the experiments with the augmentation of back-translated data to the original benchmark and our masked benchmark. We use the data\footnote{https://data.statmt.org/rsennrich/wmt16\_backtranslations} published by \citet{sennrich-etal-2016-improving}. Surprisingly, our masked benchmarks could also harness the effectiveness of back-translation. The performance improvement is equal to the original baseline on RU$\rightarrow$EN, and even amplified from $1.4$ to $1.9$ BLEU on DE$\rightarrow$EN. The results show that our masking and back-translation data focus on each different factor during train time, and both of them are also complementary.

\begin{table}[]
\centering
\begin{tabular}{lccl}
\toprule
\multirow{2}{*}{Trained on}       & \multicolumn{1}{l}{\multirow{2}{*}{DE-EN}} &
\multicolumn{2}{l}{\multirow{2}{*}{RU-EN}} \\
                             & \multicolumn{1}{l}{}                       & \multicolumn{2}{l}{}                       \\
\midrule                            
\multicolumn{1}{l}{Original} &          30.0                        & \multicolumn{2}{c}{32.2}                  \\
+ back translation                       & 31.4                                      & \multicolumn{2}{c}{33.3}                   \\

\midrule
Masked                       & 31.0                                       & \multicolumn{2}{c}{33.2}                   \\
+ back translation                       & \textbf{32.9}                                       & \multicolumn{2}{c}{\textbf{34.2}} \\
\bottomrule
\end{tabular}

\caption{
BLEU score for models trained with additional back-translation data. \textbf{Masked} : benchmark with our masking strategy whose performance in Table~\ref{tab:main_result} is the best. 
}
\label{tab:back_translation}

\end{table}

\subsection{Ablation of Content Words}

\begin{table}[]
\centering
\resizebox{\columnwidth}{!}{%
\begin{tabular}{clcc}
\toprule
                       & Trained on  & \# sents  & Accuracy \\
\midrule                    
\multirow{2}{*}{DE$\rightarrow$EN} & w/o content  & 8,997,570 & \textbf{86.2}     \\
                       & \hfill $\llcorner$ Small  & 24,040 & 79.9     \\
\midrule                       
\multirow{2}{*}{RU$\rightarrow$EN} & w/o content & 4,069,180 & \textbf{81.1}     \\
                       & \hfill $\llcorner$ Small & 24,040    & 78.6          \\
\midrule                       
\multirow{1}{*}{newstest} & w/o content & 24,040 & 68.9          \\
\bottomrule                       
\end{tabular}%
}

\caption{
Accuracy of corpus classifier trained on the data with masking on its content words. The composition of corpus is equal to the setting in Table~\ref{tab:corpus_clf}.
}
\label{tab:ablation}

\end{table}
\begin{table*}[]
\centering
\begin{tabular}{cccccc}
\toprule
\multicolumn{3}{c}{DE-EN}     & \multicolumn{3}{c}{RU-EN}        \\ \midrule
phrase     & $I$    & Mask Ratio & phrase & $I$        & Mask Ratio \\ \midrule
    . - ( NOUN    & 2.63 & 98.2\%       & you by NOUN -    & 3.41 & 0.9\%       \\
the NOUN ' s    & 2.63 & 43.6\%       & them to us . & 3.41   & 0.1\%       \\
$[$ $[$ NOUN NOUN  & 2.62 & 17.4 \%       & as ADJ as ADJ    & 3.40 & 0.2\%       \\
I should VERB to & 2.61 & 5.1\%       & Above you can VERB   & 3.40 &   0.1\%       \\
When would you VERB & 2.60 & 0.00\%       & to you by NOUN    & 3.40 & 0.4\%       \\ \midrule
–    & 1.43 & 99.9\%     & –   & 1.54       & 99.6\%       \\
‘    & 1.89 & 99.9\%       &    ‘    & 0.98    & 99.6\%       \\
“    & 1.13 & 98.7\%       &    ”    &  2.02    & 98.2\%       \\
« & 0.55 & 98.5\%       &    “   & 1.96 & 96.5\%       \\ 
» & 0.95 & 97.5\%       &     «   &   2.71    & 95.8\%      \\ \bottomrule
\end{tabular}

\caption{
Comparison between the frequently appeared phrases and \textit{fluency noise} in our best model. \textbf{Mask Ratio} : the percentage of how many tokens in given phrase pattern is used as \textit{fluency noise} in our masked benchmark. POS tag is ignored for calculation. Upper the mid-line, we sorted 4-gram phrases by its importance score $I$, defined in Equation~\ref{equation:ngram}. Below, we sorted uni-gram phrases by its Mask Ratio.
}
\label{tab:ngram}
\end{table*}
There could be a concern that the classification model is biased on the content words, and the function words would not have much influence on its accuracy. We can't deny that content words are important factors to distinguish sentences, however we also prove that the model can predict the source of sentence with only the function words to some extent.

Table~\ref{tab:ablation} shows the accuracy of classifiers trained with masking on content words of training data. We also report the result on the shrunk set to compare with newstest, too. Despite the lower performance than the result in Table~\ref{tab:corpus_clf}, we could notice that our corpus classifier also benefits from the syntactic order of function words or their positions with the accuracy of 81.1-86.2\%. Interestingly, the performance drop on newstest is the biggest compared to Table~\ref{tab:corpus_clf}, meaning that the evidence of classification in newstest is relatively concentrated on the semantic of sentence, not its structure.

\subsection{Fluency Noise}
\label{sssec:fleuncy_noise}

In this section, we discuss about \textit{fluency noise} to visually analyze what tokens disturb the training of translation. We demonstrate the specifically frequent n-gram phrases in each translation benchmark, and confirm which phrases are substantially selected as \textit{fluency noise}.

We may define \textit{fluency noise} as several patterns appearing in training corpus. Then can we catch the patterns with only heuristic statistics? We prepare the answer by using simple frequent-based method to detect the unique and specific patterns in each benchmark. Therefore, we define the importance score of given n-gram phrase $p$ in corpus $C$ as follows :

\begin{equation}
\label{equation:ngram}
    \begin{aligned}
    P(p \mid C) = {\# p \text{ } in\text{ }C \over \sum_{p' \in C} \# p' \text{ }in\text{ }C} \\
    I(p \mid C) = {P(p \mid C) \over P(p)}
    \end{aligned}
\end{equation}

For given all benchmarks, we can get the probability of $P(p \mid C)$ by taking the sum of the counts of all n-gram $p'$ in C as the denominator. We also get the prior of n-gram feature without the consideration of corpus, and this would be $P(p)$. We define the importance score $I(p \mid C)$ by just dividing the n-gram probability of given corpus by its prior. The term is similar to Pointwise Mutual Information (PMI), but we don't consider the size of each corpus (e.g. $P(C)$) since we set them equal.

\begin{table}[]
\centering
\begin{tabular}{ccc}
\toprule
phrase             & From  & Mask Ratio \\ \midrule
And it VERB to     & DE-EN & 40.2\%     \\
towards a ADJ NOUN & DE-EN & 15.4\%     \\
its NOUN of NOUN   & RU-EN & 28.4\%     \\
NOUN shall be VERB & RU-EN & 43.0\%    \\ \bottomrule
\end{tabular}

\caption{
    Examples of 4-gram phrases not including special tokens. The experimental setting is the same as Table~\ref{tab:ngram}.
}
\label{tab:wo_special}
\end{table}
\begin{table*}[]
\centering
\resizebox{1.96\columnwidth}{!}{%
\begin{tabular}{l|l}
\hline
Source   & „Die Stadt ist definitiv in der Comedy-DNA von Larrys gesamter Arbeit“, sagte Steinberg. \\
Baseline & “The city is definitely in Larry’s Comedy DNA,” said Steinberg.                          \\
Ours     & "The city is definitely in the comedy DNA of Larry's entire work," said Steinberg.       \\ \hline
Source   & Nathaniel P. Morris ist Student im zweiten Jahr an der Harvard Medical School.           \\
Baseline & Nathaniel P. Morris is a student at Harvard Medical School in the second year.           \\
Ours     & Nathaniel P. Morris is a second-year student at Harvard Medical School.                  \\ \hline
\end{tabular}%
}

\caption{
Examples of translation output
in WMT-14 DE$\rightarrow$EN test set.
}
\label{tab:case_study}
\end{table*}

We extract 1M sentences from each 3 corpus : two bilingual corpus from WMT-14 DE$\rightarrow$EN \& WMT-14 RU$\rightarrow$EN, and one monolingual corpus from News Crawl 2014. We select uni-gram and 4-gram phrases to calculate the importance score respectively. For convenience to visualize, we replace the content words with its POS tag. We filtered the phrases which appeared less than 1K times. Lastly, we measure the percentage of how many times the phrases are substantially masked by our methodology, considering the entire examples as the population. The result of analysis is demonstrated in Table~\ref{tab:ngram}. To begin with the conclusion, we find that \textit{fluency noise} is not directly correlated to the frequency of phrases in given corpus. Each of top-5 results with almost same importance score has polarized chance to be masked. However, the uni-gram tokens sorted by their mask percentage consist of almost special characters including dash or left-right quotation marks. We guess this phenomenon occurs due to the monolingual prior of BERT. Since BERT is already pre-trained on huge monolingual corpus, the common English words such as \textit{would} or \textit{us} are hard to be the evidence of the classification. On the other hand, special characters like right-single quotation mark tends to be the evidence since the quotation marks are generally written as neutral in English texts. 

Of course, it doesn't mean that only special tokens are harmful for translation. In Table~\ref{tab:wo_special}, various phrases which are not including special tokens are masked with 15-40\% of ratio. Although the same pattern of specific phrase can be used to several different sentences, the awkwardness determined by our corpus classifier could differ. The point to focus on is that masking \textit{fluency noise} helps the translation model not to strictly align the function words, which may not follow the prior of monolingual corpus.

\section{Case Study}

In Table~\ref{tab:case_study}, we bring the examples that our final model gets better score than the baseline. Upper example shows the effect of automatic regularization that our model is not affected by special tokens frequently used in source language, such as low- or left- quotation marks. Another example demonstrates our model is free from the strict alignment of function words which may be uncommon in monolingual English texts.

\section{Reliability of BLEU}

BLEU \citep{papineni-etal-2002-bleu} has been widely used as the standard evaluation metric in translation task. However, the metric calls many researchers in doubt since it isn't often correlated to human evaluation \cite{mathur-etal-2020-tangled, riley-etal-2020-translationese}. There has also been studies about the replacement of current metric \cite{sellam-etal-2020-bleurt, rei-etal-2020-comet}.

Nonetheless, the request of human evaluation for each work of translation is not efficient and coherent due to its implicit standard and the variance of annotators. For that reason, we claim that we should need to consider the ability of what BLEU could measure and the uncertainty of what BLUE could not detect, respectively. The case studies in Table~\ref{tab:case_study} shows that the semantic of target samples is not much different from the source sentence's. If the purpose of evaluation is just focused on the transfer of semantic, we may give them the almost equal score. However, the final goal of machine translation is to make the output sentence as close as possible to the distribution of test benchmark. The distribution would diverge by its text domain, nationality, and other relative factors. 

Back to the beginning, it is the fact that BLEU has the lack of ability to evaluate the equality of semantic. Apart from the flaw of the metric, we should also talk about if BLUE can judge whether the generated sentence of translation model is well adapted to the test distribution. Data-centric approach would be one of the solution to suit our desire. As \citet{freitag-etal-2020-bleu} reported, the correlation between BLEU and human evaluation is enhanced by increasing the variance of the reference text in the test set. In our work, we instead regulate the train set to be similar to the monolingual prior which is closer to the test distribution. We argue that BLEU is still effective for evaluating the degree of this adaptation, such as neutral quotation marks or \textit{second-year student} in Table~\ref{tab:case_study}.

\section{Conclusion and Discussion}

In this work, we tackle the bias hurting the fluency of model in training corpus. The bias would be uncountable and differ by its source language, the purpose of translation, the nationality of speaker, etc. However, we train the classifier and automatically detect the bias of fluency comparing to monolingual texts. The extracted bias with high confidence of classifier are masked during the translation, and it is effective enough to enhance the quality of our translation model. Throughout the analysis, we prove that the bias is concentrated on the representations which may be uncommon in monoligual English texts including special characters. We also discuss about the topic that our methodology could be applied to.

\paragraph{Low-Resource Setting} As our work only targets rich-resource benchmarks, we can conduct the same experiment on low-resource setting. Since utilizing monolingual data \citep{currey-etal-2017-copied} or language model prior \citep{baziotis-etal-2020-language} to regularize the low-resource translation model has been worked, our method can also follow the prior knowledge of them. However, our corpus classifier needs enough training data to distinguish and it would be the main challenge whether the classifier can properly be trained on the small amount of low-resource benchmarks.

\paragraph{Domain Adaptation} As the studies of using out-of-domain data in translation task have been worked \cite{aharoni-goldberg-2020-unsupervised,  muller-etal-2020-domain}, our method would be used to find the domain bias in augmented train sentences. Since \textit{fluency noise} is only defined in the range of function words against the monolingual data, we could expand it to find the tokens which are uncommon in the target domain. If the range of definition starts to include the content words, preventing the hallucination problem would be the main concern to solve.

\section*{Acknowledgements}

This research was supported by (1)
National Research Foundation of Korea (NRF-
2020R1A2C3010638), (2) the MSIT (Ministry of
Science and ICT), Korea, under the ICT Creative
Consilience program (IITP-2021-2020-0-01819)
supervised by the IITP (Institute for Information \&
communications Technology Planning \& Evalua-
tion).

\bibliography{reference}

\begin{thebibliography}{37}
\expandafter\ifx\csname natexlab\endcsname\relax\def\natexlab#1{#1}\fi

\bibitem[{Aharoni and Goldberg(2020)}]{aharoni-goldberg-2020-unsupervised}
Roee Aharoni and Yoav Goldberg. 2020.
\newblock Unsupervised domain clusters in pretrained language models.
\newblock In \emph{Proceedings of the 58th Annual Meeting of the Association
  for Computational Linguistics}, pages 7747--7763.

\bibitem[{Baker et~al.(1993)Baker, Francis, and Tognini-Bonelli}]{baker}
Mona Baker, Gill Francis, and Elena Tognini-Bonelli. 1993.
\newblock Corpus linguistics and translation studies: Implications and
  applications.
\newblock \emph{Text and Technology: In Honour of John Sinclair}, pages
  233--250.

\bibitem[{Ba{\~n}{\'o}n et~al.(2020)Ba{\~n}{\'o}n, Chen, Haddow, Heafield,
  Hoang, Espl{\`a}-Gomis, Forcada, Kamran, Kirefu, Koehn, Ortiz~Rojas,
  Pla~Sempere, Ram{\'\i}rez-S{\'a}nchez, Sarr{\'\i}as, Strelec, Thompson,
  Waites, Wiggins, and Zaragoza}]{banon-etal-2020-paracrawl}
Marta Ba{\~n}{\'o}n, Pinzhen Chen, Barry Haddow, Kenneth Heafield, Hieu Hoang,
  Miquel Espl{\`a}-Gomis, Mikel~L. Forcada, Amir Kamran, Faheem Kirefu, Philipp
  Koehn, Sergio Ortiz~Rojas, Leopoldo Pla~Sempere, Gema
  Ram{\'\i}rez-S{\'a}nchez, Elsa Sarr{\'\i}as, Marek Strelec, Brian Thompson,
  William Waites, Dion Wiggins, and Jaume Zaragoza. 2020.
\newblock {P}ara{C}rawl: Web-scale acquisition of parallel corpora.
\newblock In \emph{Proceedings of the 58th Annual Meeting of the Association
  for Computational Linguistics}, pages 4555--4567.

\bibitem[{Baziotis et~al.(2020)Baziotis, Haddow, and
  Birch}]{baziotis-etal-2020-language}
Christos Baziotis, Barry Haddow, and Alexandra Birch. 2020.
\newblock Language model prior for low-resource neural machine translation.
\newblock In \emph{Proceedings of the 2020 Conference on Empirical Methods in
  Natural Language Processing (EMNLP)}, pages 7622--7634.

\bibitem[{Binder et~al.(2016)Binder, Montavon, Bach, M{\"{u}}ller, and
  Samek}]{lrpuse}
Alexander Binder, Gr{\'{e}}goire Montavon, Sebastian Bach, Klaus{-}Robert
  M{\"{u}}ller, and Wojciech Samek. 2016.
\newblock Layer-wise relevance propagation for neural networks with local
  renormalization layers.
\newblock In \emph{International Conference on Artificial Neural Networks},
  pages 63--71.

\bibitem[{Cai et~al.(2021)Cai, Wang, Li, Lam, and Liu}]{cai-etal-2021-neural}
Deng Cai, Yan Wang, Huayang Li, Wai Lam, and Lemao Liu. 2021.
\newblock Neural machine translation with monolingual translation memory.
\newblock In \emph{Proceedings of the 59th Annual Meeting of the Association
  for Computational Linguistics and the 11th International Joint Conference on
  Natural Language Processing (Volume 1: Long Papers)}, pages 7307--7318.

\bibitem[{Caswell et~al.(2019)Caswell, Chelba, and
  Grangier}]{caswell-etal-2019-tagged}
Isaac Caswell, Ciprian Chelba, and David Grangier. 2019.
\newblock Tagged back-translation.
\newblock In \emph{Proceedings of the Fourth Conference on Machine Translation
  (Volume 1: Research Papers)}, pages 53--63.

\bibitem[{Cettolo et~al.(2012)Cettolo, Girardi, and
  Federico}]{cettolo-etal-2012-wit3}
Mauro Cettolo, Christian Girardi, and Marcello Federico. 2012.
\newblock {WIT}3: Web inventory of transcribed and translated talks.
\newblock In \emph{Proceedings of the 16th Annual conference of the European
  Association for Machine Translation}, pages 261--268.

\bibitem[{Clark et~al.(2021)Clark, August, Serrano, Haduong, Gururangan, and
  Smith}]{clark-etal-2021-thats}
Elizabeth Clark, Tal August, Sofia Serrano, Nikita Haduong, Suchin Gururangan,
  and Noah~A. Smith. 2021.
\newblock All that{'}s {`}human{'} is not gold: Evaluating human evaluation of
  generated text.
\newblock In \emph{Proceedings of the 59th Annual Meeting of the Association
  for Computational Linguistics and the 11th International Joint Conference on
  Natural Language Processing (Volume 1: Long Papers)}, pages 7282--7296.

\bibitem[{Currey et~al.(2017)Currey, Miceli~Barone, and
  Heafield}]{currey-etal-2017-copied}
Anna Currey, Antonio~Valerio Miceli~Barone, and Kenneth Heafield. 2017.
\newblock Copied monolingual data improves low-resource neural machine
  translation.
\newblock In \emph{Proceedings of the Second Conference on Machine
  Translation}, pages 148--156.

\bibitem[{Devlin et~al.(2019)Devlin, Chang, Lee, and
  Toutanova}]{devlin-etal-2019-bert}
Jacob Devlin, Ming-Wei Chang, Kenton Lee, and Kristina Toutanova. 2019.
\newblock {BERT}: Pre-training of deep bidirectional transformers for language
  understanding.
\newblock In \emph{Proceedings of the 2019 Conference of the North {A}merican
  Chapter of the Association for Computational Linguistics: Human Language
  Technologies, Volume 1 (Long and Short Papers)}, pages 4171--4186.

\bibitem[{Edunov et~al.(2018)Edunov, Ott, Auli, and
  Grangier}]{edunov-etal-2018-understanding}
Sergey Edunov, Myle Ott, Michael Auli, and David Grangier. 2018.
\newblock Understanding back-translation at scale.
\newblock In \emph{Proceedings of the 2018 Conference on Empirical Methods in
  Natural Language Processing}, pages 489--500.

\bibitem[{Freitag et~al.(2020)Freitag, Grangier, and
  Caswell}]{freitag-etal-2020-bleu}
Markus Freitag, David Grangier, and Isaac Caswell. 2020.
\newblock {BLEU} might be guilty but references are not innocent.
\newblock In \emph{Proceedings of the 2020 Conference on Empirical Methods in
  Natural Language Processing (EMNLP)}, pages 61--71.

\bibitem[{Goutte et~al.(2012)Goutte, Carpuat, and
  Foster}]{goutte-etal-2012-impact}
Cyril Goutte, Marine Carpuat, and George Foster. 2012.
\newblock The impact of sentence alignment errors on phrase-based machine
  translation performance.
\newblock In \emph{Proceedings of the 10th Conference of the Association for
  Machine Translation in the Americas: Research Papers}.

\bibitem[{Hoang et~al.(2018)Hoang, Koehn, Haffari, and
  Cohn}]{hoang-etal-2018-iterative}
Vu~Cong~Duy Hoang, Philipp Koehn, Gholamreza Haffari, and Trevor Cohn. 2018.
\newblock Iterative back-translation for neural machine translation.
\newblock In \emph{Proceedings of the 2nd Workshop on Neural Machine
  Translation and Generation}, pages 18--24.

\bibitem[{Koehn(2005)}]{europarl-2005}
Philipp Koehn. 2005.
\newblock Europarl: A parallel corpus for statistical machine translation.
\newblock In \emph{Conference Proceedings: the tenth Machine Translation
  Summit}, pages 79--86.

\bibitem[{Koppel and Ordan(2011)}]{koppel-ordan-2011-translationese}
Moshe Koppel and Noam Ordan. 2011.
\newblock Translationese and its dialects.
\newblock In \emph{Proceedings of the 49th Annual Meeting of the Association
  for Computational Linguistics: Human Language Technologies}, pages
  1318--1326.

\bibitem[{Kurokawa et~al.(2009)Kurokawa, Goutte, and
  Isabelle}]{kurokawa-etal-2009-automatic}
David Kurokawa, Cyril Goutte, and Pierre Isabelle. 2009.
\newblock Automatic detection of translated text and its impact on machine
  translation.
\newblock In \emph{Proceedings of Machine Translation Summit XII: Papers}.

\bibitem[{Li et~al.(2016)Li, Chen, Hovy, and
  Jurafsky}]{li-etal-2016-visualizing}
Jiwei Li, Xinlei Chen, Eduard Hovy, and Dan Jurafsky. 2016.
\newblock Visualizing and understanding neural models in {NLP}.
\newblock In \emph{Proceedings of the 2016 Conference of the North {A}merican
  Chapter of the Association for Computational Linguistics: Human Language
  Technologies}, pages 681--691.

\bibitem[{Liu et~al.(2020)Liu, Gu, Goyal, Li, Edunov, Ghazvininejad, Lewis, and
  Zettlemoyer}]{liu-etal-2020-multilingual-denoising}
Yinhan Liu, Jiatao Gu, Naman Goyal, Xian Li, Sergey Edunov, Marjan
  Ghazvininejad, Mike Lewis, and Luke Zettlemoyer. 2020.
\newblock Multilingual denoising pre-training for neural machine translation.
\newblock \emph{Transactions of the Association for Computational Linguistics},
  8:726--742.

\bibitem[{Liu et~al.(2019)Liu, Ott, Goyal, Du, Joshi, Chen, Levy, Lewis,
  Zettlemoyer, and Stoyanov}]{liu2019roberta}
Yinhan Liu, Myle Ott, Naman Goyal, Jingfei Du, Mandar Joshi, Danqi Chen, Omer
  Levy, Mike Lewis, Luke Zettlemoyer, and Veselin Stoyanov. 2019.
\newblock Roberta: A robustly optimized bert pretraining approach.
\newblock \emph{arXiv preprint arXiv:1907.11692}.

\bibitem[{Mathur et~al.(2020)Mathur, Baldwin, and
  Cohn}]{mathur-etal-2020-tangled}
Nitika Mathur, Timothy Baldwin, and Trevor Cohn. 2020.
\newblock Tangled up in {BLEU}: Reevaluating the evaluation of automatic
  machine translation evaluation metrics.
\newblock In \emph{Proceedings of the 58th Annual Meeting of the Association
  for Computational Linguistics}, pages 4984--4997.

\bibitem[{Miao et~al.(2021)Miao, Meng, Liu, Zhou, and
  Zhou}]{miao-etal-2021-prevent}
Mengqi Miao, Fandong Meng, Yijin Liu, Xiao-Hua Zhou, and Jie Zhou. 2021.
\newblock Prevent the language model from being overconfident in neural machine
  translation.
\newblock In \emph{Proceedings of the 59th Annual Meeting of the Association
  for Computational Linguistics and the 11th International Joint Conference on
  Natural Language Processing (Volume 1: Long Papers)}, pages 3456--3468.

\bibitem[{M{\"u}ller et~al.(2020)M{\"u}ller, Rios, and
  Sennrich}]{muller-etal-2020-domain}
Mathias M{\"u}ller, Annette Rios, and Rico Sennrich. 2020.
\newblock Domain robustness in neural machine translation.
\newblock In \emph{Proceedings of the 14th Conference of the Association for
  Machine Translation in the Americas (Volume 1: Research Track)}, pages
  151--164.

\bibitem[{Papineni et~al.(2002)Papineni, Roukos, Ward, and
  Zhu}]{papineni-etal-2002-bleu}
Kishore Papineni, Salim Roukos, Todd Ward, and Wei-Jing Zhu. 2002.
\newblock {B}leu: a method for automatic evaluation of machine translation.
\newblock In \emph{Proceedings of the 40th Annual Meeting of the Association
  for Computational Linguistics}, pages 311--318.

\bibitem[{Post(2018)}]{post-2018-call}
Matt Post. 2018.
\newblock A call for clarity in reporting {BLEU} scores.
\newblock In \emph{Proceedings of the Third Conference on Machine Translation:
  Research Papers}, pages 186--191.

\bibitem[{Pylypenko et~al.(2021)Pylypenko, Amponsah-Kaakyire, Dutta~Chowdhury,
  van Genabith, and Espa{\~n}a-Bonet}]{pylypenko-etal-2021-comparing}
Daria Pylypenko, Kwabena Amponsah-Kaakyire, Koel Dutta~Chowdhury, Josef van
  Genabith, and Cristina Espa{\~n}a-Bonet. 2021.
\newblock Comparing feature-engineering and feature-learning approaches for
  multilingual translationese classification.
\newblock In \emph{Proceedings of the 2021 Conference on Empirical Methods in
  Natural Language Processing}, pages 8596--8611.

\bibitem[{Raffel et~al.(2019)Raffel, Shazeer, Roberts, Lee, Narang, Matena,
  Zhou, Li, and Liu}]{2019-t5}
Colin Raffel, Noam Shazeer, Adam Roberts, Katherine Lee, Sharan Narang, Michael
  Matena, Yanqi Zhou, Wei Li, and Peter~J. Liu. 2019.
\newblock Exploring the limits of transfer learning with a unified
  text-to-text.
\newblock \emph{arXiv preprint arXiv:1910.10683}.

\bibitem[{Raunak et~al.(2021)Raunak, Menezes, and
  Junczys-Dowmunt}]{raunak-etal-2021-curious}
Vikas Raunak, Arul Menezes, and Marcin Junczys-Dowmunt. 2021.
\newblock The curious case of hallucinations in neural machine translation.
\newblock In \emph{Proceedings of the 2021 Conference of the North American
  Chapter of the Association for Computational Linguistics: Human Language
  Technologies}, pages 1172--1183.

\bibitem[{Rei et~al.(2020)Rei, Stewart, Farinha, and
  Lavie}]{rei-etal-2020-comet}
Ricardo Rei, Craig Stewart, Ana~C Farinha, and Alon Lavie. 2020.
\newblock {COMET}: A neural framework for {MT} evaluation.
\newblock In \emph{Proceedings of the 2020 Conference on Empirical Methods in
  Natural Language Processing (EMNLP)}, pages 2685--2702.

\bibitem[{Riley et~al.(2020)Riley, Caswell, Freitag, and
  Grangier}]{riley-etal-2020-translationese}
Parker Riley, Isaac Caswell, Markus Freitag, and David Grangier. 2020.
\newblock Translationese as a language in {``}multilingual{''} {NMT}.
\newblock In \emph{Proceedings of the 58th Annual Meeting of the Association
  for Computational Linguistics}, pages 7737--7746.

\bibitem[{Sellam et~al.(2020)Sellam, Das, and Parikh}]{sellam-etal-2020-bleurt}
Thibault Sellam, Dipanjan Das, and Ankur Parikh. 2020.
\newblock {BLEURT}: Learning robust metrics for text generation.
\newblock In \emph{Proceedings of the 58th Annual Meeting of the Association
  for Computational Linguistics}, pages 7881--7892.

\bibitem[{Sennrich et~al.(2016{\natexlab{a}})Sennrich, Haddow, and
  Birch}]{sennrich-etal-2016-improving}
Rico Sennrich, Barry Haddow, and Alexandra Birch. 2016{\natexlab{a}}.
\newblock Improving neural machine translation models with monolingual data.
\newblock In \emph{Proceedings of the 54th Annual Meeting of the Association
  for Computational Linguistics (Volume 1: Long Papers)}, pages 86--96.

\bibitem[{Sennrich et~al.(2016{\natexlab{b}})Sennrich, Haddow, and
  Birch}]{sennrich-etal-2016-neural}
Rico Sennrich, Barry Haddow, and Alexandra Birch. 2016{\natexlab{b}}.
\newblock Neural machine translation of rare words with subword units.
\newblock In \emph{Proceedings of the 54th Annual Meeting of the Association
  for Computational Linguistics (Volume 1: Long Papers)}, pages 1715--1725.

\bibitem[{Vaswani et~al.(2017)Vaswani, Shazeer, Parmar, Uszkoreit, Jones,
  Gomez, Łukasz Kaiser, and Polosukhin}]{vaswani}
Ashish Vaswani, Noam Shazeer, Niki Parmar, Jakob Uszkoreit, Llion Jones,
  Aidan~N Gomez, Łukasz Kaiser, and Illia Polosukhin. 2017.
\newblock Attention is all you need.
\newblock In \emph{Advances in Neural Information Processing Systems}, pages
  5998--6008.

\bibitem[{Voita et~al.(2019)Voita, Talbot, Moiseev, Sennrich, and
  Titov}]{voita-etal-2019-analyzing}
Elena Voita, David Talbot, Fedor Moiseev, Rico Sennrich, and Ivan Titov. 2019.
\newblock Analyzing multi-head self-attention: Specialized heads do the heavy
  lifting, the rest can be pruned.
\newblock In \emph{Proceedings of the 57th Annual Meeting of the Association
  for Computational Linguistics}, pages 5797--5808.

\bibitem[{Wu and Ong(2021)}]{seq_clf}
Zhengxuan Wu and Desmond~C. Ong. 2021.
\newblock On explaining your explanations of {BERT:} an empirical study with
  sequence classification.
\newblock \emph{arXiv preprint arXiv:2101.00196}.

\end{thebibliography}
\bibliographystyle{acl_natbib}

\end{document}